\documentclass[twoside, lettersize, journal]{IEEEtran} 
\usepackage{amsmath,amsfonts}
\usepackage{algorithmic}
\usepackage{algorithm}
\usepackage{array}
\usepackage[caption=false,font=normalsize,labelfont=sf,textfont=sf]{subfig}
\usepackage{textcomp}
\usepackage{stfloats}
\usepackage{url}
\usepackage{verbatim}
\usepackage{graphicx}
\usepackage{xcolor}
\usepackage{cite}
\usepackage{multirow}
\usepackage[normalem]{ulem}
\useunder{\uline}{\ul}{}
\usepackage{booktabs}

\begin{document}

\title{Multi-scale attention-based instance segmentation for measuring crystals with large size variation}

\author{Theresa~Neubauer, Astrid~Berg, Maria~Wimmer, Dimitrios~Lenis, David~Major, Philip~Matthias~Winter, Gaia~Romana~De~Paolis, Johannes~Novotny, Daniel~Lüftner, Katja~Reinharter and Katja~Bühler

\thanks{
    	VRVis is funded by BMK, BMDW, Styria, SFG, Tyrol and Vienna Business Agency in the scope of COMET - Competence Centers for Excellent Technologies (879730) which is managed by FFG.}

\thanks{T.~Neubauer, A. Berg, M. Wimmer, D. Lenis, D. Major, P. Winter, G.~de~Paolis, J. Novotny and K. Bühler are with VRVis Zentrum für Virtual Reality und Visualisierung Forschungs-GmbH, 1220 Vienna, Austria.}
\thanks{D.~Lüftner and K.~Reinharter are with RHI Magnesita GmbH, 8700 Leoben, Austria.}
}%

\markboth{IEEE TRANSACTIONS ON INSTRUMENTATION AND MEASUREMENT, 2023}%
{Neubauer \MakeLowercase{\textit{et al.}}: Multi-scale attention instance segmentation for large crystal size variation}

\IEEEpubid{\footnotesize{
		\begin{tabular}[t]{@{}c@{}}\copyright2023 IEEE. Personal use is permitted, but republication/redistribution requires IEEE permission. See https://www.ieee.org/publications/rights/index.html for more information.
			\\This article has been accepted for publication in IEEE Transactions on Instrumentation and Measurement. \\ Citation information: DOI 10.1109/TIM.2023.3345916 
	    \end{tabular} } }

\maketitle

\begin{abstract}
Quantitative measurement of crystals in high-resolution images allows for important insights into underlying material characteristics. Deep learning has shown great progress in vision-based automatic crystal size measurement, but current instance segmentation methods reach their limits with images that have large variation in crystal size or hard to detect crystal boundaries.
Even small image segmentation errors, such as incorrectly fused or separated segments, can significantly lower the accuracy of the measured results.
Instead of improving the existing pixel-wise boundary segmentation methods, we propose to use an instance-based segmentation method, which gives more robust segmentation results to improve measurement accuracy.
Our novel method enhances flow maps with a size-aware multi-scale attention module.
The attention module adaptively fuses information from multiple scales and focuses on the most relevant scale for each segmented image area. 
We demonstrate that our proposed attention fusion strategy outperforms state-of-the-art instance and boundary segmentation methods, as well as simple average fusion of multi-scale predictions.
We evaluate our method on a refractory raw material dataset of high-resolution images with large variation in crystal size and show that our model can be used to calculate the crystal size more accurately than existing methods.

\end{abstract}

\begin{IEEEkeywords}
Crystal size measurement, grain size, microscopic high-resolution images, refractory raw material dataset, attention model, deep learning.

\end{IEEEkeywords}

\section{Introduction}
\begin{figure}[!t]
	\centering
	\includegraphics[width=1\columnwidth]{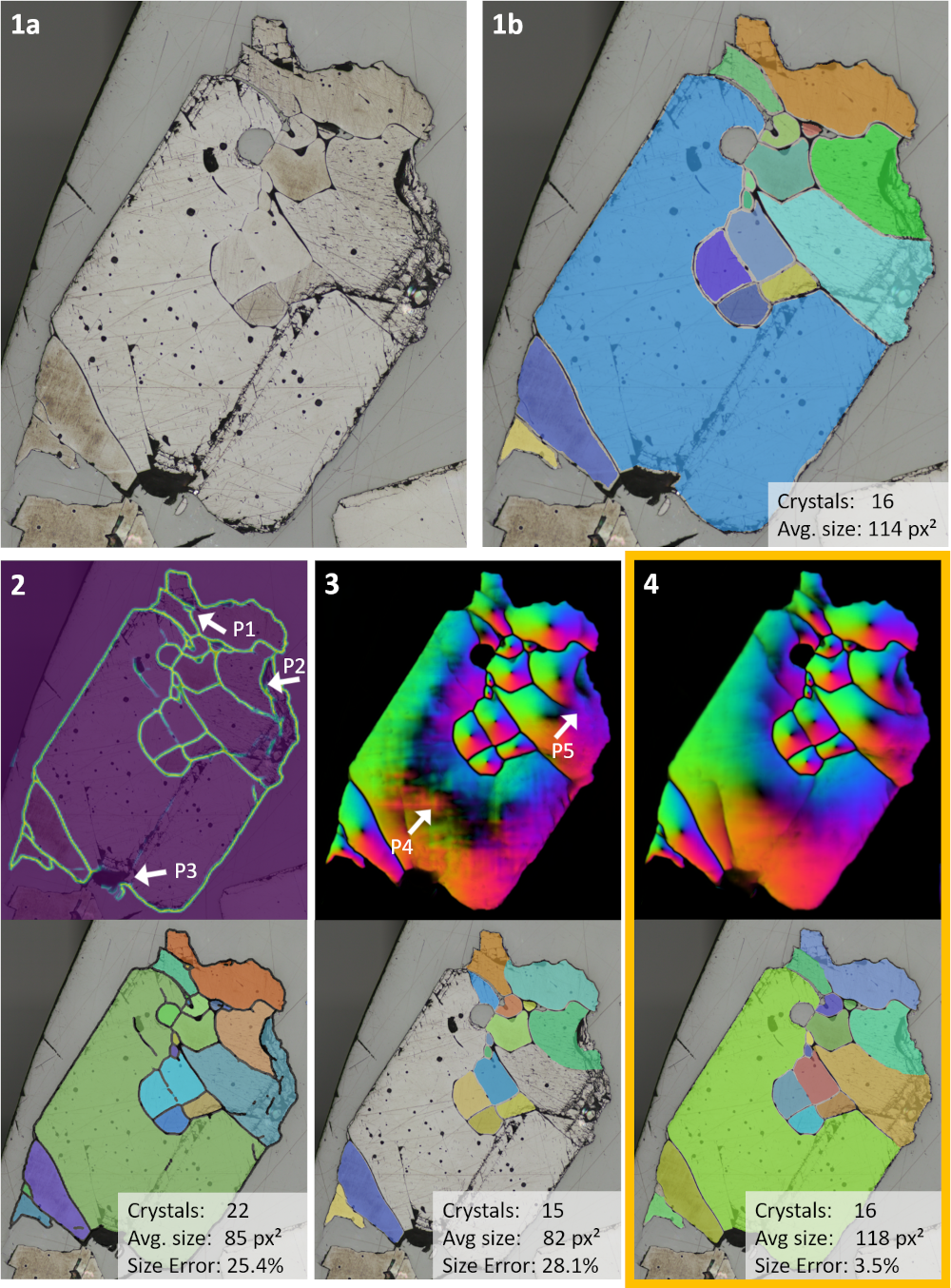}	
	\caption{\textbf{Comparison of different segmentation methods and their impact on size measurement accuracy.} \textbf{(1a)} Grain with multiple crystals of different sizes. \textbf{(1b)}~Ground truth crystal instance segmentation.
	\textbf{(2)} Boundary prediction and resulting instance segmentation from  Yu et al. \cite{yuCenterGuidedConnectivityPreservingNetwork2022}.
	The boundary model mispredicts tiny crystals (P1) when the crystal boundary is difficult to identify. Also scratched surfaces (P2) are falsely recognized as boundary. 
	Gaps (P3) between grain and background need to be closed by post-processing.
	\textbf{(3)} Intermediate flow map output and final instance segmentation from the instance model (Cellpose) of Stringer et al. \cite{stringerCellposeGeneralistAlgorithm2021}. 
	The model cannot correctly identify the center of the large crystal (P4), and it also fails to detect thin boundaries (P5), resulting in missing instance segments. \textbf{(4)}~Our proposed multi-scale attention model overcomes these problems and correctly segments crystals of varying sizes and thus enhances the accuracy of the measured sizes.}
	\label{fig1}
\end{figure}

\IEEEPARstart{M}{easuring} the size of crystals or grains in minerals on high-resolution microscopic images is an important task for various fields such as mineralogy, material science, and geoscience. 
The average crystal size is a key metric in the quantitative analysis and provides information about underlying material properties. 
In recent years, image analysis applications have increasingly been used to support this analysis in numerous domains such as material science, medicine or biology \cite{choudharyRecentAdvancesApplications2022, litjensSurveyDeepLearning2017}. The automation of vision-based measurement has the major advantages of providing more accurate results, faster measurement time, reduction of manual preparation steps, and improved reproducibility \cite{shirmohammadiCameraInstrumentRising2014}. 
Depending on the type of material, the appearance of the crystals and their boundaries varies. 
\IEEEpubidadjcol 
In general, a crystal is an area with a uniform color and a contrasting boundary line separating it from adjacent crystals. One way to measure crystal sizes is through classic image processing methods like intensity value thresholding or edge detection \cite{peregrina-barretoAutomaticGrainSize2013} to detect the borders of crystals and measure the properties of resulting enclosed regions. However, in practice, these methods quickly reach their limits when the data becomes more complex.
Challenges include barely visible boundary lines between adjacent crystals, scratches caused by polishing that are misidentified as boundary lines, broken crystal regions that do not allow a clear delimitation due to high-contrast areas, or the existence of impurities, pores or noise in the image (see Figure~\ref{fig1}). The most common problem with classical image processing methods is the over-segmentation of boundary lines, where too many edges in the image are recognized as boundaries. Another challenge is that very fine boundaries may not be detected or may result in fragmented segmentation lines. 
Both of these detection errors result in incorrectly separated or fused segments, which can severely degrade the accuracy of size and count measurements.
In recent studies \cite{choudharyRecentAdvancesApplications2022}, deep learning-based segmentation methods are getting more popular as they are able to learn the context-dependent appearance of complex data. In the case of material science on high resolution microscopic images of polycrystalline samples, also known as micrographs, deep learning is ideally suited to handle the varying appearance of crystals and boundaries and has been shown to improve segmentation results.

The state-of-the-art (SOTA) approach to measure crystal size is to use instance segmentation \cite{stringerCellposeGeneralistAlgorithm2021} first and then calculate statistics on size distribution of the resulting segments.
However, as we show, instance segmentation faces difficulties when images have (1) large variations in crystal sizes, and (2) a lot of crystals that are packed together, separated by very thin boundary lines. 
In this paper we address these challenges of instance and boundary segmentation by proposing a novel method that combines instance segmentation with size-aware multi-scale attention (SiMA).

We adapt a SOTA instance segmentation model for microscopic images called Cellpose \cite{stringerCellposeGeneralistAlgorithm2021}, and enhance it with a size-aware attention model, inspired by crowd size estimation methods on natural images \cite{jiangAttentionScalingCrowd2020}. 
Our framework can overcome the large size variation issue of instance segmentation by adaptively fusing information from multiple scales and focus on the most relevant regions for segmentation at each scale, as shown in Figure \ref{fig2} in Section \ref{method}. 

Our aim is to calculate the crystal sizes more accurately by improving the crystal segmentation performance. We focus on difficult images, containing adjacent crystals of varying sizes, which can neither be solved by simple resizing nor by tiling strategies. We evaluate our method on a dataset of refractory raw material samples, which have a polycrystalline structure and large variations in crystal size.

We summarize our main contributions:
\begin{itemize}
	\item  We introduce SiMA, a novel module that enables dense instance segmentation on high-resolution images to improve crystal size measurement with large size variation.
	
	\item SiMA is an attention module that improves the fusion of multi-scale instance predictions. We combine flow maps from multi-scale inputs with a size-aware attention module. For the multi-scale fusion process, SiMA emphasizes results at the optimal image resolution for the area to be segmented, while properly handling conflicting information from other scales.
	
	\item Our resource-efficient method improves crystal segmentation over the state-of-the-art on a very challenging dataset, while having much less parameters than state-of-the-art transformer-based instance segmentation models. We demonstrate that especially on difficult cases our method is capable of improving the crystal size measurement accuracy on images with instances of varying sizes. We significantly outperform the state-of-the-art methods in single-scale and boundary-based crystal segmentation.
\end{itemize}

\section{Related Work}
\noindent 
From the literature on grain and crystal size measurement, we investigate three main approaches for the initial image segmentation method: boundary segmentation (\ref{Boundary_segmentation_method}), superpixel segmentation (\ref{superpixel}), and instance segmentation (\ref{instance_proposal}, \ref{instance_proposal_free}). We also review multi-scale segmentation methods and attention models for segmentation tasks (\ref{multi_scale}).

\subsection{Boundary segmentation method} \label{Boundary_segmentation_method}
\noindent The  most popular method for grain/crystal segmentation is boundary segmentation, where a binary semantic segmentation task is performed. 
The boundary pixels are segmented as foreground and non-boundary pixels are segmented as background. 
In a post-processing step, all pixels located within the same boundary are assigned to the same label, resulting in an enumeration of separated instances, an \textit{instance label map}. 
To perform the boundary segmentation task, the works in this field are using neural networks with U-Net inspired architectures, e.g., \cite{yuCenterGuidedConnectivityPreservingNetwork2022, bachmannEfficientReconstructionPrior2022, liGrainBoundaryDetection2020a, dasDeepNeuralNetworks2022}.
The boundary method has a significant drawback, since neighboring crystals are fused together if the segmented boundary mistakenly has gaps. 
This, in turn, leads to incorrect measurement of the number and size of crystals. 
Therefore, post-processing strategies have been proposed to close fragmented segmentation boundaries, e.g., rule-based morphological operations like dilation and erosion \cite{banerjeeAutomatedMethodologyGrain2019, podorSEraMicSemiautomaticMethod2021}, applying the watershed algorithm to complete partially segmented crystal boundaries \cite{bachmannEfficientReconstructionPrior2022}, boundary skeletonization to automatically delete or extend discontinued boundaries \cite{liAutomationInterceptMethod2022} or GAN-based approaches to automatically close gaps in areas where the boundary is occluded by impurities \cite{liGrainBoundaryDetection2020a}.
However, these post-processing strategies are not capable of closing all boundary gaps and may also incorrectly connect boundary lines. 
Some deep learning papers are thus focusing on topology-preserving models to avoid predictions with fragmented boundary lines in the first place. 
Yu et al. \cite{yuCenterGuidedConnectivityPreservingNetwork2022} train a model with centerline dice loss \cite{shitClDiceNovelTopologyPreserving2021} as part of their loss function. Bachmann et al. \cite{bachmannEfficientReconstructionPrior2022} propose a weighted cross-entropy and Jaccard loss to tackle the imbalanced foreground/background classes. 
Liu et al. \cite{liuBoundaryLearningUsing2022} predict an adaptive boundary-weighted map based on the original U-Net distance transform \cite{ronnebergerUNetConvolutionalNetworks2015}. 
While these proposed methods encourage deep learning models to focus on topology preservation during training, there is no guarantee that these models will actually predicting topology correctly. In difficult cases, such as occluded borders or vanishing boundaries, there could be multiple correct solutions to draw the segmentation boundary, but only one (random) solution is given in the annotation mask in the training process. 

\subsection{Superpixel segmentation} \label{superpixel}
\noindent To overcome the boundary segmentation issues, another approach is to segment the crystal regions instead of the boundaries \cite{maitreMineralGrainsRecognition2019a, yuSuperpixelSegmentationsThin2023, dabekSuperpixelBasedGrainSegmentation2023}, utilizing super-pixel segmentation to cluster pixels of the same color. These pixel clusters represent the final instance segmentation map. The superpixel segmentation overcomes the problem with vanishing boundaries as long as the adjacent crystals have different colors, but would fuse adjacent crystals with the same color. Moreover, it falsely divides the crystal into different clusters when the crystal has different color areas. Another disadvantage is that background areas or impurities are also segmented as individual instances. 

\subsection{Instance segmentation – Proposal-based} \label{instance_proposal}
\noindent The third approach for instance segmentation is proposal-based instance segmentation, also known as ‘detect-and-segment’ approach. It is a two-stage method, where first potential candidate regions are extracted and in a next step the binary segmentation of the proposed regions is predicted. The most prominent representative of this group is Mask \mbox{R-CNN} \cite{heMaskRCNN2017}. 
By now, several variants of Mask R-CNN have been developed, including RTMDet-Ins \cite{lyuRTMDet2022}, which is designed to segment instances in real-time.
In mineralogy and metallography, Mask R-CNN or variants \cite{kohUtilisingConvolutionalNeural2021, liPriorMaskRCNN2021} have been applied to perform instance segmentation of grains. 
However, the grains in the demonstration datasets usually have large distances to each other and touch each other only in rare cases.
Studies \cite{stringerCellposeGeneralistAlgorithm2021, shiImprovedUNetImage2022} with datasets of densely packed instances with no gaps between the instances, often report that Mask R-CNN has lower evaluation scores compared to their own methods. 
Brebandere et al. \cite{debrabandereSemanticInstanceSegmentation2017} state that proposal-based instance segmentation methods have difficulties to extract the segmentation mask if the proposed region contains more than one instance. 

Bhukarev et al. \cite{bukharevTaskInstanceSegmentation2018} propose a different concept for instance segmentation of densely packed mineral grains of sandstone. In the first step, region proposals are predicted using a Fully Convolutional Network (FCN) \cite{longFullyConcolutional2015} to extract the grain centers from the image. The image is then split into regions around the grain centers. For each image crop, a segmentation mask is predicted using a segmentation model. The segmented masks are then merged into the final instance segmentation map. This approach is capable of segmenting densely packed regions, but has the drawback that these small image crops may lack the global context to perform accurate segmentation. Also, the task of cropping and predicting each instance reduces time performance when there are many small instances in an image.

\subsection{Instance segmentation – Proposal-free} \label{instance_proposal_free}
\noindent Previous papers \cite{debrabandereSemanticInstanceSegmentation2017, uhrigPixelLevelEncodingDepth2016, liangProposalFreeNetworkInstanceLevel2018a} on proposal-free approaches present models that learn pixel embeddings, such that pixels belonging to the same instance have similar embeddings and pixels from different instances have dissimilar embeddings. The resulting pixel feature vectors are then post-processed with a clustering algorithm to create instance segmentations. 
While effective, the proposed clustering methods are slow and also the number of clusters is not known in advance.
Recent transformer-based architectures, such as MaskFormer \cite{ChengSK21} and OneFormer \cite{Jain0C0OS23}, speed up the instance segmentation task by using the pixel decoder embeddings and an additional transformer decoder to generate binary masks for each object query. However, the downside is that they are heavy on parameters and computation, and are limited by the maximum number of instances per sample.

Another approach known as center-regression is to train a deep learning model where each pixel is assigned a 2D-vector pointing towards the instance center \cite{uhrigBox2PixSingleShotInstance2018, kendallMultitaskLearningUsing2018, nevenInstanceSegmentationJointly2019a, chengPanopticDeepLabSimpleStrong2020a}. The two-channel output map consists of the horizontal and vertical pixel offsets (distance) to its centers. In a post-processing step, each pixel is shifted by the predicted offset and assigned to the closest center. This results in the final instance segmentation map, whereby pixels with the same assigned center have the same label. With the exception of \cite{aguilar3DFiberSegmentation2021}, which uses the center-regression approach in material science, these approaches were mainly applied to natural images.

\textit{Cellpose: }
Cellpose \cite{stringerCellposeGeneralistAlgorithm2021}, a proposal-free instance segmentation method, was originally developed for cell segmentation, but Stringer et al. \cite{stringerCellposeGeneralistAlgorithm2021} show that Cellpose generalizes well to other types of densely packed objects. The network architecture is a U-Net variant which creates a center-regression flow map, consisting of a horizontal and a vertical gradient map. 
Instead of a simple coordinate offset, the flow map is a vector field of normalized gradients produced by a heat diffusion from the median center. An additional output is the pixel probability map of the foreground. 
In a post-processing step, Euler integration is used to construct the flow from each pixel to the center to finally produce the instance segmentation. 
Due to the large size of the micrograph, usually, the model is applied to each patch and the results are stitched, since strong downsampling of the image is not possible due to fine structures and small objects that would no longer be visible. Previous stitching approaches deal with binary segmentation masks leading to mistakenly fused objects or open boundaries. Cellpose overcomes these problems: due to the characteristics of the flow field and the mask assignment with the Euler integration, flow patches can be easily stitched or combined. 

Our approach utilizes Cellpose's strategy of combining patches of flow maps and extends it to fuse multiple predictions from different image scales. 

\subsection{Multi-scale segmentation models and Attention models} \label{multi_scale}
\noindent In semantic segmentation with FCNs, the issue of scale variations is approached by utilizing multi-scale features \cite{chenAttentionScaleScaleaware2016, chenDeepLabSemanticImage2018}. In the context of instance segmentation, only some works are focusing on multi-scale approaches, e.g., on natural images \cite{liuPathAggregationNetwork2018} or remote sensing images \cite{chengCrossScaleFeatureFusion2021}.Chen et al. \cite{chenAttentionScaleScaleaware2016} propose an attention map to combine image features at different scales. For each scale, the attention model produces a pixel-wise weight map that has the largest weights for objects that belong to that particular scale. 
One use-case involving severe scale variations of objects is crowd size estimation of people, where  intermediate density maps are used to distinguish between different levels of crowdedness to improve the count estimation \cite{liuADCrowdNetAttentionInjectiveDeformable2019, xuAutoScaleLearningScale2022, jiangAttentionScalingCrowd2020}. Jiang et al. \cite{jiangAttentionScalingCrowd2020} propose a density attention network that produces a binary segmentation map for each predefined density level. Their crowd size estimation framework includes an attention model called Density Attention Network (DANet) that produces for each density level a binary segmentation map. To obtain these density level maps, they blur the annotation of heads of people with a Gaussian kernel and use a threshold set to divide the calculated density map into multiple density levels.\\

Inspired by the work of Chen et al. \cite{chenAttentionScaleScaleaware2016} and Jiang et al.~\cite{jiangAttentionScalingCrowd2020} on scale-aware attention maps, we develop SiMA, a model for size-aware attention maps. 
We combine SiMA with Cellpose \cite{stringerCellposeGeneralistAlgorithm2021}, a SOTA method in instance segmentation,  to obtain improved prediction results, benefitting from different scale levels. 
Tiling strategies used in previous works may result in a lack of global context and may negatively impact the model's segmentation performance. To avoid tiling the image, the image size can also be scaled to the input size of the model. On one hand, downsizing the image before patch creation retains a portion of the overall context. On the other hand, heavy downsampling may result in the omission of small crystals. 
However, using a predefined input resolution can lead to poor segmentation results, as some crystals are too small or too large to be segmented correctly. 
To our knowledge we are the first to suggest attention-based fusion as a way to counter this problem and improve crystal/grain segmentation over the SOTA in instance segmentation as well as boundary segmentation.

\section{Method} \label{method}
\begin{figure*}[!t]
	\includegraphics[width=\textwidth]{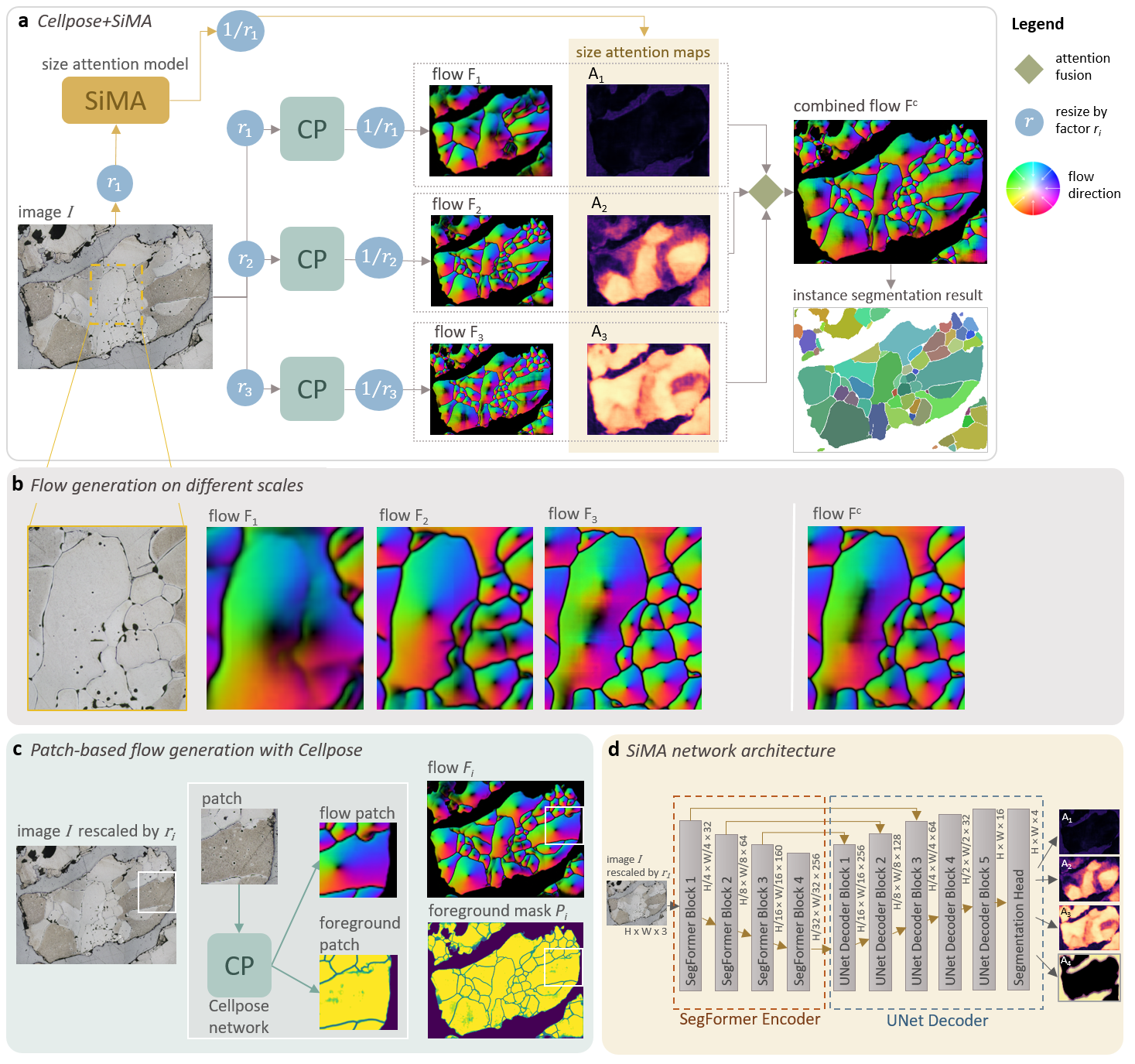}	
	\caption{\textbf{(a)} Proposed pipeline Cellpose+SiMA, visualized for $N=3$. At training time, the CP module is trained with different scaling augmentations. At inference time, the input image is rescaled and passed through Cellpose $N$ times. At different input resolutions, the flow emphasizes different details. The combination of flows via attention maps helps to produce predictions adapted to different parts of the image with varying characteristics. \textbf{(b)} The magnified image region illustrates the effect of different image resolutions, from low to high. Flow $F_{1}$ shows that important details are lost when the image resolution is too low. Flow $F_{2}$ has an ideal image resolution to detect the left big crystal, but has problems with tiny crystals. The center of the big crystal is not detected correctly in flow $F_{3}$, indicating that the image resolution is too high to fit the big crystal on one patch as model input.
	The combined flow $F^{c}$ shows the correct region-wise fusion of the flow maps using the weights from the attention maps of SiMA. \textbf{(c)} At test time, the rescaled image is tiled into patches, which are then processed by Cellpose. An advanced stitching algorithm merges the output patches together.
	\textbf{(d)} Network architecture of the SiMA model. }
	\label{fig2}
\end{figure*}

\noindent
As shown in Figure \ref{fig2}, our proposed architecture consists of two major components: (1) a model for size-aware attention (SiMA) and (2) an instance segmentation model (CP) based on the Cellpose architecture \cite{stringerCellposeGeneralistAlgorithm2021}. 
To effectively segment crystals of different crystal sizes within one image, our architecture aims to distribute the attention on different crystal sizes into different output layers.
A size-aware architecture addresses scale issues in micrographs, since crystals usually exhibit large size differences, where some regions contain densely packed crystals while other regions contain only a single large crystal.

\subsection{Cellpose}
\noindent We base our instance segmentation model on the architecture of Cellpose \cite{stringerCellposeGeneralistAlgorithm2021}, which is a U-Net variant that creates a flow map and a foreground probability map. We change the input layer of the Cellpose architecture from grayscale to RGB input. 
For the training procedure, we use manually segmented ground truth instance labels to create flow maps and foreground maps. 

\noindent The original Cellpose paper tries to solve the problem of large size differences between objects by using image scaling as a preprocessing step. 
Images with large objects in the ground truth masks are scaled down and images with small objects are scaled up until the average object size in each image matches a target size $d$. 
This way, the model can learn to predict images within a specific size range. 
Further image augmentation also includes additional random scaling, translation, and rotation.
After augmentation, a random patch of size $s^\mathit{CP}$ is cropped from the image and the corresponding foreground mask $P$ and flow map $F$ and used for training.
At inference time, the average object size of an image is not known, so Cellpose uses a built-in SizeModel to estimate this quantity. 
The SizeModel is a regression model that takes as input the extracted feature maps of the Cellpose encoder that were generated from an image patch. 
To resize the image to an ideal input resolution, the scale factor is computed by matching the estimated size to the predefined target size $d$. 
The resized image is then tiled in overlapping patches of size $s^{CP}$. 
Cellpose is applied to each patch and the resulting flow patches and foreground patches are stitched together. 
To create a smooth transition between  patches, the average of the overlapping regions is calculated using a sigmoidal taper mask at the patch borders to mitigate the impact of border predictions. The reconstructed image is then resized to the original image resolution.

\subsection{Cellpose+SiMA}
\noindent The described scale preprocessing of Cellpose does not solve the crystal size variation problem. Since crystals with large size differences can be found next to each other, there is usually no single ideal input resolution per image. To address this problem, we enhance a Cellpose model by our size-aware attention module SiMA.
We first define $N$ resize factors $r_{i}$, $i \in \{1, ...,N\}$, by which the original input image $I$ is resized, resulting in $N$ different input image resolutions in ascending order, $(r_{1}, ..., r_{N}) $.
Here, the first resize factor $r_1$ resizes the image to the input resolution of the neural network of Cellpose and the last resize factor $r_N$ is 1.0 and maintains the original resolution of the image,  while $r_2$ to $r_{N-1}$ are intermediate resolutions between those two. 

For each $r_{i}$, the resized image is passed through the instance segmentation model CP. 
From CP we obtain a flow map $F_{i}$ pointing pixels to the crystal center and a foreground probability map $P_{i}$, which are then resized by factor $\frac{1}{r_{i}}$  to the original image resolution.
If the scaled image is larger than the input size $s^{CP}$ of CP, the image is tiled in overlapping patches of size $s^{CP}$ and the predictions are stitched using the method given in \cite{stringerCellposeGeneralistAlgorithm2021}, illustrated in Figure \ref{fig2}(c).

A simple approach to combine predictions from multiple input resolution flow maps is to calculate the average or maximum of all predictions. However, we argue that the impact of the different multi-scale predictions should be adjusted region-wise according to the crystal size through an advanced region-wise merging procedure, inspired by \cite{jiangAttentionScalingCrowd2020}. Image regions with small crystals should receive increased attention on the prediction with the highest input resolution, medium crystals should be predicted at medium input resolution, and large crystals at low resolution, i.e., they should be downsampled to benefit from increased context. 

\subsubsection{SiMA}
\begin{figure}[!t]
	\centering
	\includegraphics[width=0.75\columnwidth]{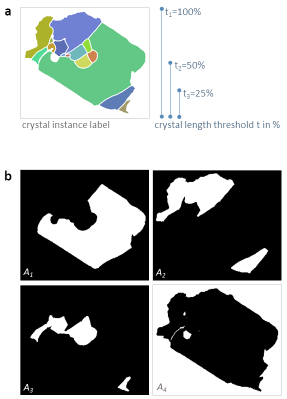}
	\caption{\textbf{(a)} Crystal instance label with \textbf{(b)} the corresponding attention maps for $N=3$ and $t=(100\%, 50\%, 25\%)$. Each attention map is a binary segmentation of  all crystal instances belonging to a certain length range of $t$. }
	\label{fig3}
\end{figure}

We adopt the idea of binary attention segmentation maps for crowd size estimation from \cite{jiangAttentionScalingCrowd2020} and propose our new size-aware attention model called SiMA. SiMA is a multi-class segmentation model, based on a U-Net variant, where each output layer represents an attention map, which is a binary segmentation mask.
To this end, SiMA aims to represent regions of different crystal sizes by providing $N+1$ input image-sized weight maps called attention maps $A=(A_{1},..., A_{N+1})$ (see Figure \ref{fig3} for an example attention map). $A_{N+1}$ is the background mask, where pixels are set to 1 if no crystal is present. 
To preserve the global context of the image, the image is resized to the model's input size $s^\text{SiMA}$. The predicted attention maps are then resized back with resize factor $\frac{1}{r_{1}}$ to the original image resolution.
In contrast to DANet from Jiang et al. \cite{jiangMethodAutomaticGrain2018a}, where segmentation masks for different density levels by blurring center pixels were obtained, the ground truth for our attention maps $A_{i}$ is directly created from the instance label map by applying a crystal length threshold $t_{i}$ for each size level. For this purpose, the crystal length is calculated by using the maximum of height and width from an axis-aligned bounding box around the crystal’s segmentation label.
The threshold values $t=(t_{1}, ..., t_{N})$ are given as a percentage of the image size in decreasing order. 
The first attention map represents all crystals with lengths between the threshold values $t_{1}$ and $t_{2}$.  We continue with this thresholding procedure until the remaining crystals with lengths between $t_{N}$ and 0 are assigned to the last attention map $A_{N}$. Finally, pixels belonging to crystals of the same size level, given by $t_{i}$ and $t_{i+1}$, form the binary segmentation mask of attention map $A_{i}$. The background pixels are collected in an additional attention map $A_{N+1}$. The final output is produced by performing sigmoid operation across all attention maps. 

\subsubsection{Multi-scale attention fusion with Cellpose+SiMA}
To calculate the combined flow map $F^{c}$, we compute the pixel-wise weighted average of the flow maps $F_{1}$ to $F_{N}$ using the weights $A_{i}$ for each $F_{i}$. For the combined foreground mask $P^{c}$ we also calculate the weighted average of $P_{1}$ to $P_{N}$ using the corresponding weights $A_{i}$ for each $P_{i}$. To calculate the final instance segmentation, we map the flow field $F^{c}$ to an instance segmentation using the gradient flow tracking method from \cite{stringerCellposeGeneralistAlgorithm2021}. Only pixels of $F^{c}$ where the pixel value of $P^{c}$ is above a threshold $h$ are considered. 
For each pixel, the gradient flow tracking method iteratively follows the spatial derivatives until they converge to the crystal center by Euler integration. Pixels with the same assigned center receive the same instance label, resulting in the final instance segmentation map.

\section{Experimental Evaluation}
\subsection{Dataset}
\noindent
Our dataset consists of high-resolution micrographs of polished and acid-treated refractory raw material samples taken with an optical microscope. Refractory raw material forms clusters of crystallites. These clusters (grains) can be either single crystals or polycrystals (see Figure \ref{fig1} for an example grain). Crystals of polycrystals normally have random crystallographic orientations and the transitions from one crystal to the next, called boundary, can be seen as dark line or color changes on micrographs. As part of the microstructure assessment, mineralogy experts first polish and etch the material sample to make its crystalline properties observable under the microscope.
The dataset consists of 313 image cutouts of micrographs captured at different image resolutions (ranging from 170 to 270 pixels per millimeter), with each cutout cropped to one  grain. The image cutouts have varying heights and widths spanning from 150 to 1800 pixels. A grain consists of one single crystal or multiple crystals of different sizes ranging from 8 to 1750 pixels.
The labeling of the data was done by mineralogy experts. Each crystal was manually segmented, with a boundary line of about 5 pixels between adjacent segments. These labels were then converted to instance segmentation labels.

\textit{Dataset split: }
We use stratified random sampling to ensure a balanced split of grains with different characteristics and crystal size variability.
Therefore, each grain is assigned to one of three classes, representing different levels of challenges. To this end, we define a crystal size homogeneity score as the ratio between the largest and the smallest crystal for all crystals within a grain. The value ranges from 0 to 1, with values close to 1 indicating a uniform distribution of crystal sizes and values close to 0 indicating high variability of crystal sizes. Class 1 is the class of easy examples, where the maximum crystal size is smaller than the input size $s^{CP}$ of the model and therefore the image can be predicted patch-based at the original input resolution with Cellpose. 
The rest of the images are divided into class 2, consisting of images with homogeneity score smaller than 0.1 and class 3 of homogeneity score greater than 0.1. 
Class 2 represents the images where large crystals are at least 10 times larger than the smallest crystals. 
This means that rescaling the image such that the largest crystal is fully contained in the input size of the model would potentially harm the detection of smaller crystals, as they may not be detectable in case of a too large resizing factor.
Class 3 represents images with smaller variations in crystal size than in class 2, where the ideal resize factor should not cause small crystals to disappear.
The 313 samples are split into a training, a validation and a test set (60\%, 20\%, 20\%) using the defined three classes for stratification. 

\subsection{Experimental setup}
\noindent
We perform a grid search over selected hyperparameters to find the best configuration for each model (including SOTA models). Model selection is performed on the validation set of our dataset according to best PQ score (cf. Section \ref{evaluation metrics}). The searched hyperparameters and their value ranges are given at the end of each model section.

\noindent \textbf{Our framework Cellpose+SiMA:} 
CP model and SiMA have the same input size $s^{CP} =$ $ s^{\text{SiMA}}=224$ pixels.
We use $N=4$ resize factors $r_{i}$ with $r_{1}=s^{\text{SiMA}}/\max(\mathit{image\: height, image\: width})$, i.e., we scale the image to the input resolution of the neural network, $r_{2}=0.5$, $r_{3}=0.75$, and $r_{4}=1.0$. If input images are already at the CP input size $s^{CP}$, we set $r_{1}~=0.25$ to ensure a proper size hierarchy from smallest to largest.
All models are trained using the PyTorch framework. For training, we utilize the Adam optimizer and implement a cosine learning rate decay with a 10-epoch warmup learning rate that linearly increases from zero to the initial learning rate. 
If other adjacent grains are visible in the image, they are masked in the model training.

\noindent \textit{Hyperparameter space:} $r_1~\in~\{0.1, 0.25\}$, $r_2~\in~\{0.25, 0.5\}$, $r_3 \in \{0.5, 0.75\}, r_4 \in \{1.0\}, N \in \{2,3,4\}$. 

\textbf{Cellpose:} We follow the implementation of the original Cellpose approach  \cite{stringerCellposeGeneralistAlgorithm2021} and make the following adaptions in addition to the ones described above: 
(1) The number of input channels is changed from 1 to 3 to use RGB images as input, resulting in an input model size of $224\times224\times 3$. 
(2)~The initial learning rate is set to 0.1. 
(3) The minibatch size is set to 30, due to hardware limitations.
(4) The threshold value $h$ of the foreground mask $P^{c}$ is changed to 0 to also include foreground regions with higher uncertainty.
(5) We use the suggested scaling preprocessing from Cellpose for model training, but change the target average crystal size $d$ from 30 pixels to 50 pixels. The subsequent random scaling uses a factor from 0.75 to 1.25.
(6) We use additional training augmentation by adding random brightness and contrast from the albumentations library~\cite{albumentations} to the image. 
We keep all other parameters from the original implementation.

\noindent \textit{Hyperparameter space:} $ h \in \{0.0, 0.25, 0.5\}$, $ d \in \{30,$ $ 50,$ $ 100\},$  learning rate $ \in \{0.05, 0.1, 0.2, 0.4\}$
 
\textbf{SiMA:} 
The network architecture of SiMA is a U-Net variant with an input size of $224\times 224\times3$ and an output size of $224 \times 224 \times 4$. 
We use the segmentation models framework~\cite{Iakubovskii2019} to implement the network architecture. The encoder backbone is the Mix Vision Transformer B0 from SegFormer \cite{xieSegFormerSimpleEfficient2021}, with pre-trained ImageNet weights. SiMA is trained through a pixel-wise weighted average of the cross-entropy loss for each layer. The loss weights for each channel are $(2,4,6,8,1)$, with the largest weight assigned to the attention map with the smallest crystal size to account for the unbalanced segmentation task.
We train the model for 500 epochs and use an initial learning rate of $0.01$ and a minibatch size of $30$. For training augmentation, the image resizing is composed of scaling the image size to the input patch size $s^{SiMA}$ followed by random scaling factor between $0.5$ and $1.5$. We also add the same random contrast and brightness augmentation. From the original CP model implementation, we use random translations, rotations, and cropping.
After augmentation, we use the ground truth label to create four attention maps via the threshold set $t = (100\%, 50\%, 25\%, 12.5\%)$.

\noindent \textit{Hyperparameter space:} learning rate $ \in \{0.01, 0.05, 0.1\}$, loss weights $\in \{ (2,4,6,8,1), (2,3,4,5,1), (1,1,1,1)\},$ $t = \{(100\%, 50\%, 25\%, 12.5\%), (100\%, 25\%, 12.5\%, 6.25\%),$ $ (100\%, 40\%, 20\%, 10\%)\}$,  segmentation models backbone $\in$ \{Mix Vision Transformer B0, MobileNet v2, ResNext50\}.

\vspace{0.5cm}

\noindent \textbf{Comparison with SOTA instance segmentation methods}:
We compare our model with four SOTA instance segmentation methods: the boundary model of Yu et al. \cite{yuCenterGuidedConnectivityPreservingNetwork2022}, OneFormer~\cite{Jain0C0OS23}, MaskFormer \cite{ChengSK21}, and RTMDet-Ins \cite{lyuRTMDet2022}. Yu et al.'s \cite{yuCenterGuidedConnectivityPreservingNetwork2022} boundary model is the only recent work on dense instance segmentation on a material science dataset with public code that allows reproducibility. OneFormer and MaskFormer are among the top performing methods for instance segmentation on several benchmarks. We also compare our model with RTMDet-Ins \cite{lyuRTMDet2022}, a lightweight model that has a similar number of parameters to ours and currently the best real-time instance segmentation method on the MSCOCO benchmark dataset. OneFormer and MaskFormer, on the other hand, are very large models, each with more than 100 million parameters.

\textbf{Boundary model:} 
We refer to Yu et al.'s \cite{yuCenterGuidedConnectivityPreservingNetwork2022} model as boundary model.
Their dataset is not publicly available, so we train their model on our dataset using their public code for training and inference. The model input size is $512 \times 512$ pixels. For training and evaluation, we have ground truth boundary labels with a line width of 5 pixels.
We train the model for 300 epochs with a minibatch size of $16$ and a learning rate of $0.0075$.
For training augmentation, we change the brightness/contrast settings to match the other model settings and use random image crops instead of center crops. 
At inference time, the boundary prediction is converted to a binary label using a threshold operation. A boundary width of five pixels is obtained by skeletonization and dilation operations to match the ground truth data. We find an optimal boundary threshold value of $0.2$ for this process by evaluating it against the validation set and selecting the threshold value with the highest centerline dice (clDice) score \cite{shitClDiceNovelTopologyPreserving2021}. The binary boundary label is transformed to an instance label map by assigning all pixels within the same boundary to the same instance. For this procedure, we restrict the label map to the foreground of the grain to prevent boundary gaps at the grain border from falsely merging crystal instances with the background. All other parameters and strategies for training and inference are kept to their default values. We use the clDice score to measure the performance of the topology of the boundary line.

\noindent \textit{Hyperparameter space:} learning rate $ \in \{0.001, 0.0025,$ $ 0.005, $ $ 0.0075, 0.01\}$, boundary threshold $\in \{0.1, 0.2,$ $ 0.3, 0.4,$ $ 0.5\}$.

\textbf{MaskFormer:}
We finetune the MaskFormer \cite{ChengSK21} with Swin-B backbone \cite{LiuL00W0LG21}, trained on ADE20K \cite{ZhouZPFB017}, on our dataset with random $512 \times 512$ crops as input and the same augmentations as for the other models. We trained the model for 500 epochs and used an initial learning rate of  $5\mathrm{e}{-5}$ and batch size $32$. For the post-processing,  we set the probability score threshold to keep predicted instance masks to $0.25$.

\noindent \textit{Hyperparameter space:} learning rate $\in$ $\{1\mathrm{e}{-6}$, $5\mathrm{e}{-6},$ $ 1\mathrm{e}{-5},$ $ 5\mathrm{e}{-5}, $ $ 1\mathrm{e}{-4}\}$, probability score threshold $\in \{0.0, 0.25,$ $ 0.5, 0.75\}$. 

\begin{table*}[!]
	\centering
	\caption{Comparison of multi-scale fusion methods}
	\label{tab:result_merging}
	\begin{tabular}{@{}llll@{}}
		\begin{tabular}[c]{@{}l@{}}\textbf{Resize factors}\\ \textbf{(r\textsubscript{1},..., r\textsubscript{N})}\end{tabular}
		& \textbf{Multi-scale fusion method}                           &\textbf{ PQ (\%) $\uparrow$}  & 
		\begin{tabular}[c]{@{}l@{}} \textbf{ MRE } \\ \textbf{ACS (\%) $\downarrow$} \end{tabular} \\ \midrule
		\multirow{2}{*}{(r\textsubscript{1}, 1)}            &  Average               &   68.2 ± 18        & 25.2 ± 22  \\
		&  \textit{Attention;} t=(100\%, 25\%)   & 74.0 ± 15 & 21.5 ± 33 \\ \midrule
		\multirow{2}{*}{(r\textsubscript{1}, 2 $*$ r\textsubscript{1}, 1)}   & Average           &  71.3 ± 17       & 20.4 ±21  \\
		& \textit{Attention;} t=(100\%, 50\%,  25\%)  & 75.8 ± 14 & 13.0 ± 20 \\ \midrule
		\multirow{2}{*}{(r\textsubscript{1}, 0.5, 0.75, 1)} & Average &  70.4 ± 19   & 17.8 ± 20 \\
		& \textit{Attention;} t=(100\%, 50\%,  25\%,   12.5\%)  & \textbf{77.1 ± 15} & \textbf{11.8 ± 19} \\ \midrule
	\end{tabular}
\end{table*}

\textbf{OneFormer:}
We finetune the OneFormer \cite{Jain0C0OS23} with Swin-T backbone (shi-labs\text{/}oneformer-ade20k-swin-tiny from the HuggingFace hub \cite{wolf-etal-2020-transformers}), trained on ADE20K, on our dataset with random $512\times 512$ crops as input and the same augmentations as for the other models. We trained the model for 150 epochs and used an initial learning rate of  $1\mathrm{e}{-4}$ and batch size $16$. For the post-processing, we set the probability score threshold to keep predicted instance masks to $0.75$.

\noindent \textit{Hyperparameter space:} learning rate $\in$ $\{1\mathrm{e}{-6}$, $5\mathrm{e}{-6},$ $ 1\mathrm{e}{-5}, $ $ 5\mathrm{e}{-5}, 1\mathrm{e}{-4}\}$, probability score threshold $\in \{0.0, 0.25, 0.5, 0.75\}$.

\textbf{RTMDet-Ins:}
We finetune the RTMDet-Ins \cite{lyuRTMDet2022} with the RTMDet-Ins-tiny backbone from the mmdetection framework~\cite{mmdetection}, trained on COCO,  on our dataset with random $640\times640$ crops as input and adapted our augmentation setting to their pipeline. We trained the model for 200 epochs and used an initial learning rate of $1\mathrm{e}{-5}$ and batch size $16$. Due to memory limitations during training, we set the maximum number of instances per image to $200$.
For the post-processing, we set the IOU threshold of the non-maximum suppression to $0.2$, the binary segmentation mask threshold to $0.5$, and the score threshold to $0.3$.

\noindent \textit{Hyperparameter space:} learning rate $\in$ $\{1\mathrm{e}{-5}, $ $5\mathrm{e}{-5}$, $ 1\mathrm{e}{-4}$, $5\mathrm{e}{-4}$, $1\mathrm{e}{-3}\}$, IOU threshold $\in \{0.0, 0.1, 0.2\}$, mask threshold  $\in \{0.1, $ $ 0.25, 0.5\}$, score threshold $\in \{0.1,$ $ 0.2, 0.3\}$.

\subsection{Evaluation metrics}
\label{evaluation metrics}
\noindent To evaluate the instance segmentation results, we report Panoptic Quality (PQ) \cite{kirillovPanopticSegmentation2019a} and Aggregated Jaccard Index (AJI) as defined in \cite{kumarDatasetTechniqueGeneralized2017}. 
PQ is computed by matching each predicted instance $p$ with the ground truth instance $q$ that has the highest intersection over union (IoU) score, where an IoU value greater than 0.5 is required for a match. 
The matching creates three sets: matched pairs (TP), unmatched predicted segments (FP), and unmatched ground truth segments (FN). 
PQ calculates the average IoU of matched segments and applies a penalty for unmatched segments:

$$PQ = \dfrac{\sum_{(p,q)\in TP} IoU(p,q)}{|TP| + 0.5*|FP| + 0.5*|FN|} $$

To evaluate the performance of the calculated crystal sizes, we compute the average crystal size (ACS) for each grain/cluster.
The crystal size is the crystal's diameter in pixels, which is derived from the crystal area $|A|$ (number of pixels in the segmentation) under the assumption that the crystal shape is a circle. The crystal size is calculated as $ 2 \cdot \sqrt{|A|/\pi} $. ACS\textsuperscript{GT} is the average crystal size of all crystals in the grain of the ground truth label, and ACS\textsuperscript{pred}  for the predicted label.

To report the test scores, we calculate their mean and standard deviation across all test samples for PQ, AJI, and ACS. Additionally, we compute the mean absolute crystal size error (MAE ACS) and the mean relative crystal size error (MRE ACS) between ACS\textsuperscript{GT} and ACS\textsuperscript{pred}.

\subsection{Benchmark settings}
\noindent We report the average latency (i.e. network inference and post-processing time) for all models on the test set. The inference speed is tested on an NVIDIA TITAN RTX GPU (40GB) with batch size 1.
For model training, MaskFormer and OneFormer required significantly more hardware resources and training time than the other models, and were therefore trained on an NVIDIA A100 GPU (80GB).

\section{Results}

\begin{table*}[!]
	\centering
	\caption{Comparison of our proposed method with SOTA methods}
	\label{tab:result}
	\begin{tabular}{@{}lcccrrrrrrr@{}}
		& \multicolumn{3}{c}{Model settings}  & \multicolumn{4}{c}{Metrics} & \multicolumn{3}{c}{Model parameters \& latency}  \\ 
		\cmidrule(lr){2-4} \cmidrule(lr){5-8}  \cmidrule(l){9-11}
		Models &
		\begin{tabular}[c]{@{}c@{}}image \\ rescaling\end{tabular} &
		\begin{tabular}[c]{@{}c@{}}multi-\\ scale\end{tabular} &
		\begin{tabular}[c]{@{}c@{}}fusion \\ type\end{tabular} &
		\multicolumn{1}{c}{PQ (\%) $\uparrow$} &
		\multicolumn{1}{c}{AJI  (\%) $\uparrow$} &
		\multicolumn{1}{c}{\begin{tabular}[c]{@{}c@{}}MAE \\ ACS $\downarrow$\end{tabular}} &
		\multicolumn{1}{c}{\begin{tabular}[c]{@{}c@{}}MRE \\ ACS (\%) $\downarrow$\end{tabular}} &
		\multicolumn{1}{c}{\text{\#} Params} &
		\begin{tabular}[c]{@{}c@{}}Input \\ size (px)\end{tabular} &
		\begin{tabular}[c]{@{}c@{}}Latency \\ (s) $\downarrow$ \end{tabular} \\ 
		\cmidrule(r){1-1} \cmidrule(lr){2-4} \cmidrule(l){5-8}  \cmidrule(l){9-11}
		\textit{\textbf{Cellpose+SiMA (Ours)}} & \checkmark & \checkmark & attention & \textbf{77.1 ± 15} & \textbf{84.0 ± 17} & 24.3 & 11.8 ± 19 &  12.1M       &  $224^{2}$         & 3.94 \\
		\textit{Cellpose original \cite{stringerCellposeGeneralistAlgorithm2021}}                 & \checkmark &            &           & 72.1 ± 19 & 78.8 ± 24 & 56.6 & 19.4 ± 35 &  6.6M          &  $224^{2}$         & 4.96 \\
		\textit{Cellpose multi-scale}                    & \checkmark & \checkmark & average   & 70.4 ± 19 & 71.6 ± 28 & 52.4  & 17.8 ± 20 & 6.6M           &  $224^{2}$         & 3.61\\
		\textit{Cellpose no resizing}                    &            &            &           & 55.8 ± 29 & 50.4 ± 37 & 110.4 & 33.5 ± 36 &  6.6M          &  $224^{2}$         & 1.86\\
		\textit{Boundary model \cite{yuCenterGuidedConnectivityPreservingNetwork2022}}          &            &            &           & 66.6 ± 18 & 83.0 ± 12 & 39.7 & 16.0 ± 17 &  9.7M          &  $512^{2}$         & 0.49\\
		\textit{RTMDet-Ins (tiny) \cite{lyuRTMDet2022}}          &            &            &           & 65.5 ± 19 & 77.0 ± 19 & 31.5
		& 12.4 ± 14 &  5.2M          & $640^{2}$          &0.08\\
		\textit{MaskFormer \cite{ChengSK21}}          &            &            &           & 51.7 ± 18 & 66.4 ± 23 & 61.2
		 & 21.1 ± 23 &  101.8M          & $512^{2}$          & 0.21\\
		\textit{OneFormer \cite{Jain0C0OS23}}          &            &            &           & 70.8 ± 13 & 82.3 ± 11 & \textbf{18.3}
		 & \textbf{10.1 ± 10} & 507.4M          & $512^{2}$          &0.12\\
	\end{tabular}
\end{table*}
\noindent In this section, we present the experiment results on our refractory raw material dataset. To prove the effectiveness of our model, we perform an ablation study with multi-scale fusion methods and our size-aware attention model SiMA. We compare our attention fusion approach with the original Cellpose instance segmentation method \cite{stringerCellposeGeneralistAlgorithm2021}, a SOTA boundary segmentation method for grains \cite{yuCenterGuidedConnectivityPreservingNetwork2022}, as well as current SOTA instance segmentation methods. Moreover, we evaluate the performance of multi-scale fusion on image samples with different crystal size distributions.

To test the statistical significance of our results, we use 10-fold cross validation with 10 independent test sets. 
We perform Wilcoxon signed-rank test with Bonferroni correction to evaluate significant rank differences with a significance level of $\alpha=0.05$.

\subsection{Multi-scale fusion methods}
\noindent We conduct experiments on two fronts to evaluate the impact of our size-aware attention model SiMA: (1) multi-scale inputs and (2) multi-scale fusion methods. Table \ref{tab:result_merging} summarizes our findings.
We test different numbers of resize factors $(r_ {1},...r_{N})$ with $N=2$, $N=3$, and $N=4$. The smallest resize factor $r_{1}$ (resizing the image to the input size of the model) and also the largest resize factor $r_{N}$ (using the original image resolution) are the same for each $N$. For $N=3$ and $N=4$ we add intermediate resize factors $r_{2}$ to $r_{N-1}$. 
To combine the multi-scale outputs, we use our attention fusion method and compare it with pixel-wise averaging. 
In our experiment, we train SiMA with 2, 3, and 4 attention maps (excluding the background map) using different crystal length thresholds $t$ to divide the crystals into attention maps. We show that our attention fusion method outperforms the pixel-wise averaging method for each $N$. For our attention fusion method, we can also see from Table \ref{tab:result_merging} that the PQ score and the relative size error (MRE ACS) improving with a higher $N$. 
The best PQ score of 77.1±15\% and a very low relative size error of 11.8±19\% are obtained for the model with $N=4$ resize factors and attention fusion. 
We denote the attention fusion model with $N=4$ as \textit{Cellpose+SiMA} and the average fusion model with $N=4$ as \textit{Cellpose multi-scale} for further comparison.

\subsection{Comparison with State-of-the-Art}
\noindent In Table \ref{tab:result}, we compare our multi-scale attention fusion approach (Cellpose+SiMA) with the original Cellpose architecture and SOTA instance segmentation methods. 

\subsubsection{Cellpose architecture ablation}
In the original Cellpose paper, a regression model is used to determine the best image resolution for each input image and they are resized individually before using Cellpose.  
As a baseline, we evaluate the Cellpose model without the prior resizing step and achieve a PQ score of 55.8\%, which is a drastic drop in performance compared to a PQ score of 72.1\% from the Cellpose original. The mean relative error of the crystal size improves from 33.5\% to 19.4\% if using the prior resizing step.
We examine whether a single resizing step or the fusion of multi-scale outputs would lead to better performance.
Both, the attention and the average fusion boost the performance substantially compared to the baseline. However, the average fusion yields a PQ score of 70.4\%, which is inferior to 72.1\% from the Cellpose original. This suggests that one individually adapted resizing factor is preferable to average fusion of three predefined rescaling steps.  
The attention fusion beats the average fusion and all the other models in metric PQ and AJI. It has a high PQ score of 77.1\% and a low MRE ACS of 11.8\% demonstrating high segmentation quality and measurement accuracy.
We find statistically significant differences in PQ between the attention fusion mechanism (Cellpose+SiMA) compared to the average fusion (Cellpose multi-scale) and to the single resizing (Cellpose original).

\subsubsection{Boundary model}
We also evaluated the boundary model from Yu et al. against our instance segmentation models. The boundary model achieves a relatively high clDice sore of 79.99 ± 9\%, which indicates a high overlap of crystal boundaries. 
The PQ score of 66.6\% for evaluating the crystal instances is above the baseline, but it is relatively low in relation to the other models. 
We qualitatively assess the model’s segmentation results and observe many boundary gaps in cases where the crystal boundary is either very thin or obscured by black scratches (see Figures \ref{fig4} and \ref{fig5}). 
Although the boundary overlap is high, small boundary gaps lead to fused crystal instances and therefore the accuracy of the size measurement is harmed with a mean absolute crystal size error of 39.7 and a mean relative error of 16\%.

\subsubsection{SOTA instance segmentation methods}
OneFormer, MaskFormer, and RTMDet-Ins belong to the best performing instance segmentation methods on natural images. We showed that they also perform well on our dataset, with proper training and tuning. Of these models, only OneFormer performs slightly better than our model in the size error scores (MAE ACS and MRE ACS). However, our light-weight Cellpose+SiMA method has a better overlap of predicted and ground truth instances, achieving the highest PQ and AJI scores of all models.

\subsubsection{Computational complexity}
In Table \ref{tab:result}, we compare model performance with model complexity and show that our Cellpose+SiMA has the best performance metrics while belonging to the group of parameter-efficient models.
Our pipeline Cellpose+SiMA consists of two light-weight networks, SiMA with 5.5M and Cellpose with 6.6M parameters, providing greater efficiency than, e.g., MaskFormer with 101.8M parameters. Cellpose+SiMA also achieves a faster inference time than the original Cellpose, which requires a pre-processing step for size estimation that takes 1.9 seconds. On our hardware, our model's inference time is 3.95 seconds, which includes 0.04 seconds for the SiMA network, 0.4 seconds for $N=4$ times the Cellpose network, and 3.3 seconds for the post-processing of the flow maps. 
In terms of low computational cost and high performance metrics, our proposed Cellpose+SiMA provides a good trade-off between inference time and model performance.
It also offers a good trade-off between inference time and model performance, with low computational cost and high performance metrics.

\begin{figure}[!b]
	\includegraphics[width=\linewidth]{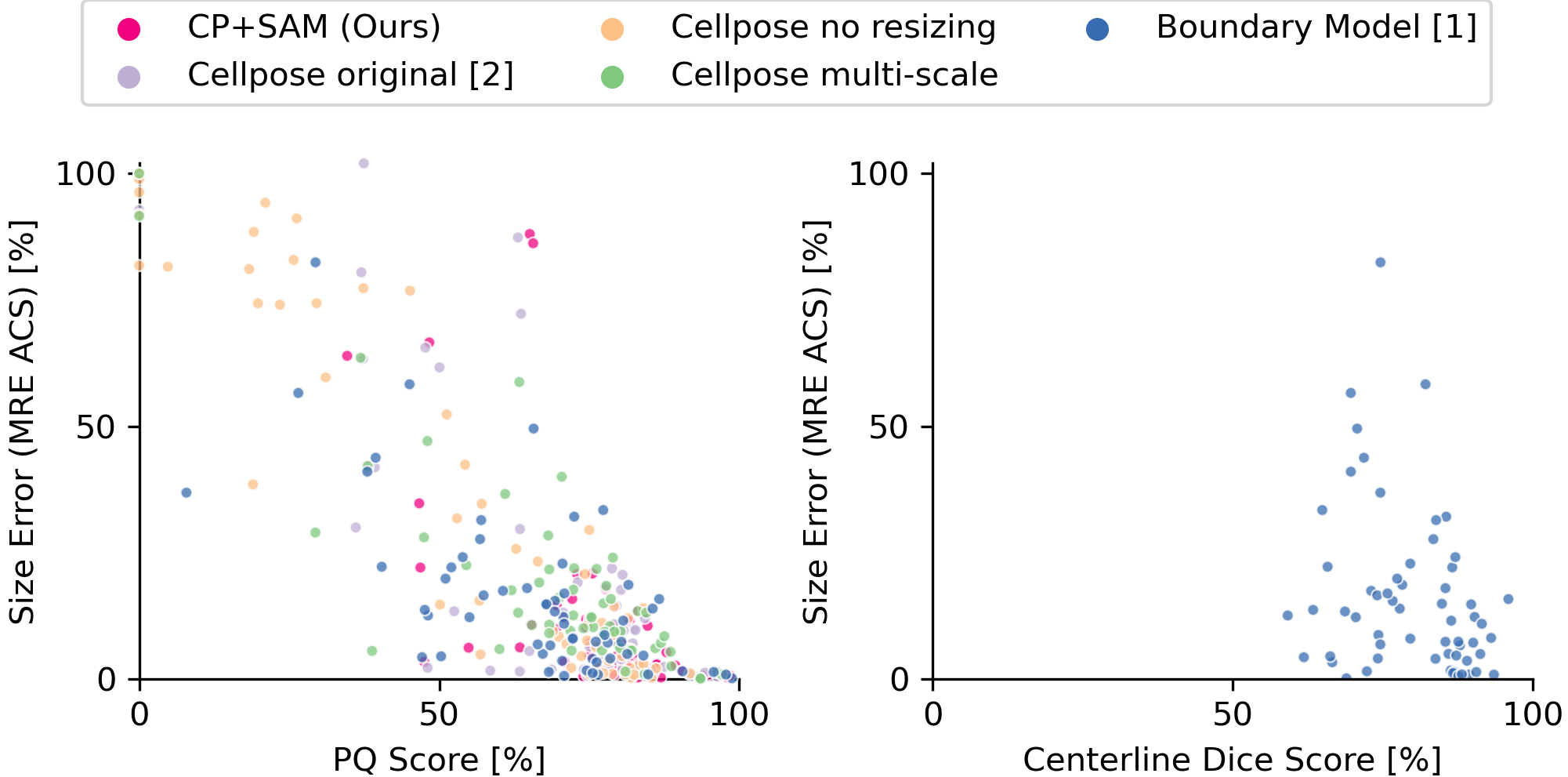}	
	\caption{The relationship between the PQ/clDice score and the relative error of the crystal size on the test set. A higher PQ score results in a lower measurement error, while a higher clDice score does not indicate more accurate size measurements. }
	\label{fig_plot}
\end{figure}

\subsection{Instance segmentation with large size variations}

\begin{table*}[!t]
	\centering
	\caption{PQ score and average crystal size results per data class}
	\label{tab:result_class}
	\begin{tabular}{@{}lrrrrrcccc@{}}
		& \multicolumn{2}{c}{PQ (\%) $\uparrow$}                & \multicolumn{3}{c}{ACS}                                      & \multicolumn{2}{c}{MAE ACS $\downarrow$}   & \multicolumn{2}{c}{MRE ACS (\%) $\downarrow$} \\ \cmidrule(lr){2-3} \cmidrule(lr){4-6} \cmidrule(lr){7-8} \cmidrule(lr){9-10}
		Class &
		\multicolumn{1}{c}{\textit{\begin{tabular}[c]{@{}c@{}}Cellpose\\ original\end{tabular}}} &
		\multicolumn{1}{c}{\textit{\begin{tabular}[c]{@{}c@{}}\textbf{Cellpose+SiMA}\\ \textbf{ (Ours)}\end{tabular}}} &
		\multicolumn{1}{c}{\textit{ACS\textsuperscript{GT}}} &
		\multicolumn{1}{c}{\textit{\begin{tabular}[c]{@{}c@{}}Cellpose\\ original\end{tabular}}} &
		\multicolumn{1}{c}{\textit{\begin{tabular}[c]{@{}c@{}}\textbf{Cellpose+SiMA}\\ \textbf{(Ours)}\end{tabular}}} &
		\multicolumn{1}{c}{\textit{\begin{tabular}[c]{@{}c@{}}Cellpose\\ original\end{tabular}}} &
		\multicolumn{1}{c}{\textit{\begin{tabular}[c]{@{}c@{}}\textbf{Cellpose+SiMA}\\ \textbf{(Ours)}\end{tabular}}} &
		\multicolumn{1}{c}{\textit{\begin{tabular}[c]{@{}c@{}}Cellpose\\ original\end{tabular}}} &
		\multicolumn{1}{c}{\textit{\begin{tabular}[c]{@{}c@{}}\textbf{Cellpose+SiMA}\\ \textbf{(Ours)}\end{tabular}}} \\ \midrule
		\multicolumn{1}{l|}{1} & \textbf{80.9 ± \hphantom{0}8}  & \multicolumn{1}{r|}{80.6 ± \hphantom{0}9}  & 72.8 ± \hphantom{0}29   & \textbf{74.3 ± \hphantom{0}29}   & \multicolumn{1}{r|}{75.8 ± \hphantom{0}29}   & \textbf{\hphantom{0}2.2}          & \multicolumn{1}{c|}{ \hphantom{0}3.3   }  &  \hphantom{0}\textbf{3.4 ±} \hphantom{0}\textbf{3} & \hphantom{0}5.0 ± \hphantom{0}4     \\
		\multicolumn{1}{l|}{2} & 69.4 ± 15 & \multicolumn{1}{r|}{\textbf{71.7 ± 15}} & 141.9 ± \hphantom{0}81  & 186.2 ± 197 & \multicolumn{1}{r|}{\textbf{171.2 ± 142}} & 70.2         & \multicolumn{1}{c|}{ \textbf{45.7}}     &  32.1 ± 50 & \textbf{22.3 ± 29}    \\
		\multicolumn{1}{l|}{3} & 66.0 ± 26 & \multicolumn{1}{r|}{\textbf{78.9 ± 18}} & 339.6 ± 192 & 313.0 ± 217 & \multicolumn{1}{r|}{\textbf{333.3 ± 194}} & 97.5         & \multicolumn{1}{c|}{ \textbf{23.9}}     &  22.8 ± 28 & \hphantom{0}\textbf{8.2 ± 10}    \\ \midrule
	\end{tabular}
\end{table*}
\noindent Our refractory raw material dataset consists of crystals with huge size differences. 
We compare the impact of the two approaches (1) single image resizing (Cellpose original) and (2) multi-scale attention fusion (Cellpose+SiMA) on the heterogeneous crystal size dataset.  
In Table \ref{tab:result_class} we evaluate the PQ and the crystal size measurement scores for each data class separately.
Class 1 contains data samples with a maximum crystal size that is smaller than the model input size. We assume that a multi-scale fusion could be redundant in this case. This assumption is supported by the slightly higher PQ score and the lower absolute/relative error scores of the Cellpose original in comparison to the Cellpose+SiMA.

Images with the largest size variation are in class 2. In this case, Cellpose+SiMA outperforms Cellpose original in PQ and MAE/MRE ACS scores. 
Our multi-scale fusion approach is capable of overcoming the challenge of crystal size variability within a single image sample and performs better than the single rescaling step of Cellpose original. 

The most significant difference between the two methods is evident in class 3, where Cellpose+SiMA has more accurate size measurements and tops Cellpose original with a higher PQ score of 78.9\% over 66.0\% and a lower relative size error of 8.2\% over 23.9\%.
Class 3 contains crystals that are larger than the input size, but have a smaller size variation than class 2. 
The weak performance of the Cellpose original on this class indicates that the single rescaling step struggles to find an ideal resizing factor, since it has difficulties estimating the crystal size of images with many large crystals. 

This confirms our assumption that region-wise multi-scale fusion is well suited for images that have crystals with large size variations. We find statistically significant differences in PQ between Cellpose+SiMA and Cellpose original on class 2 and 3.

\begin{figure*}[!t]
	\centering
	\includegraphics[width=0.95\textwidth]{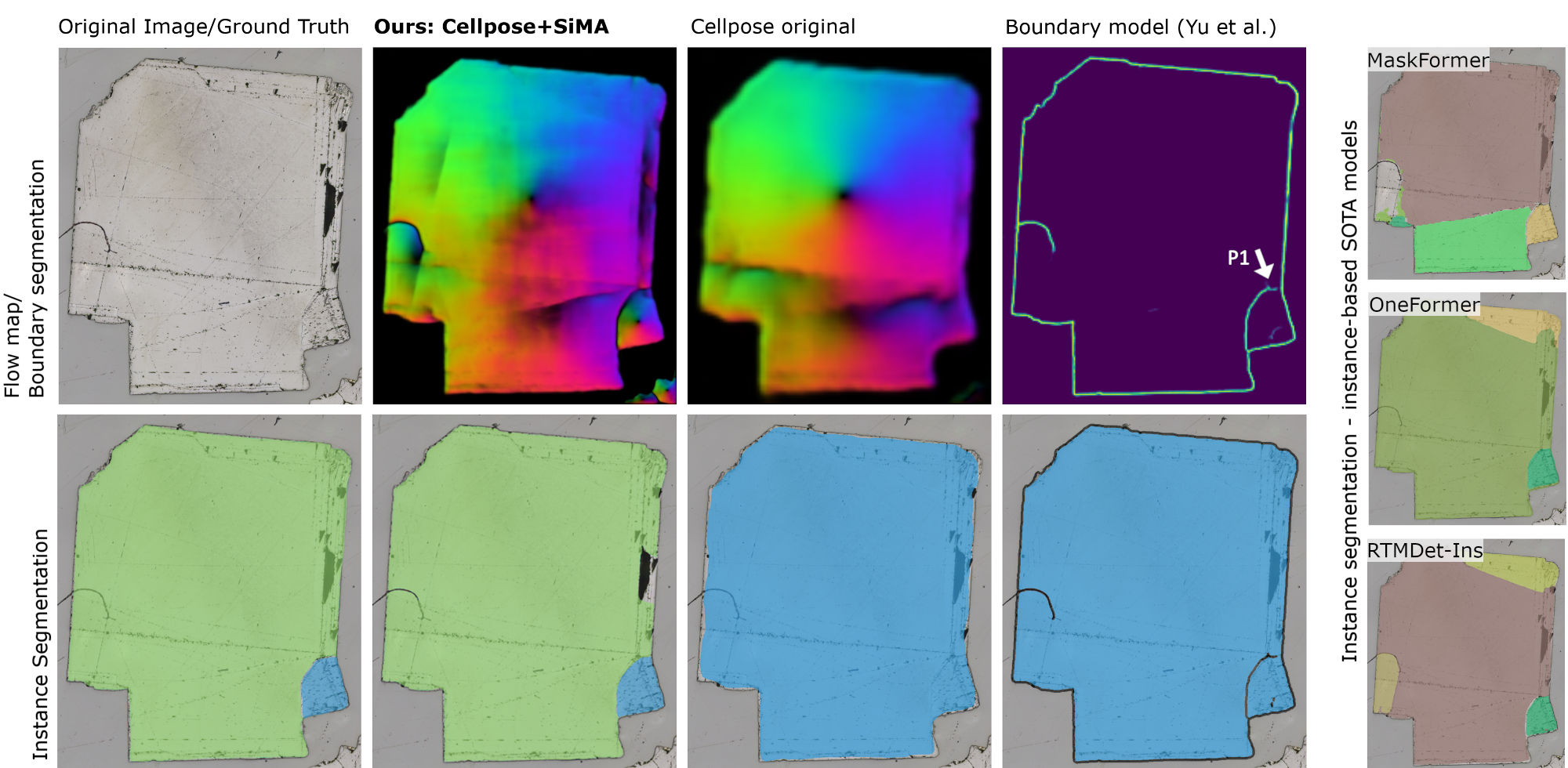}
	\caption{Visual comparison of different model results. The image shows two crystals with a large difference in size. Our Cellpose+SiMA method can accurately segment both crystals, which is enabled by our multi-scale prediction approach. Due to the strong downsampling of Cellpose original, the model is not able to recognize the thin boundary between the two crystals and the whole area is detected as one crystal. 
	The boundary prediction from Yu et al.'s model (visible as a black line in the instance segmentation) produces a boundary gap (P1) where the boundary is difficult to detect, resulting in merged crystal segments. MaskFormer, OneFormer, and RTMDet-Ins struggle to segment instances that occupy a larger area than the model's input size.}
	\label{fig4}
\end{figure*}

\begin{figure*}[!t]
	\centering
	\includegraphics[width=1\textwidth]{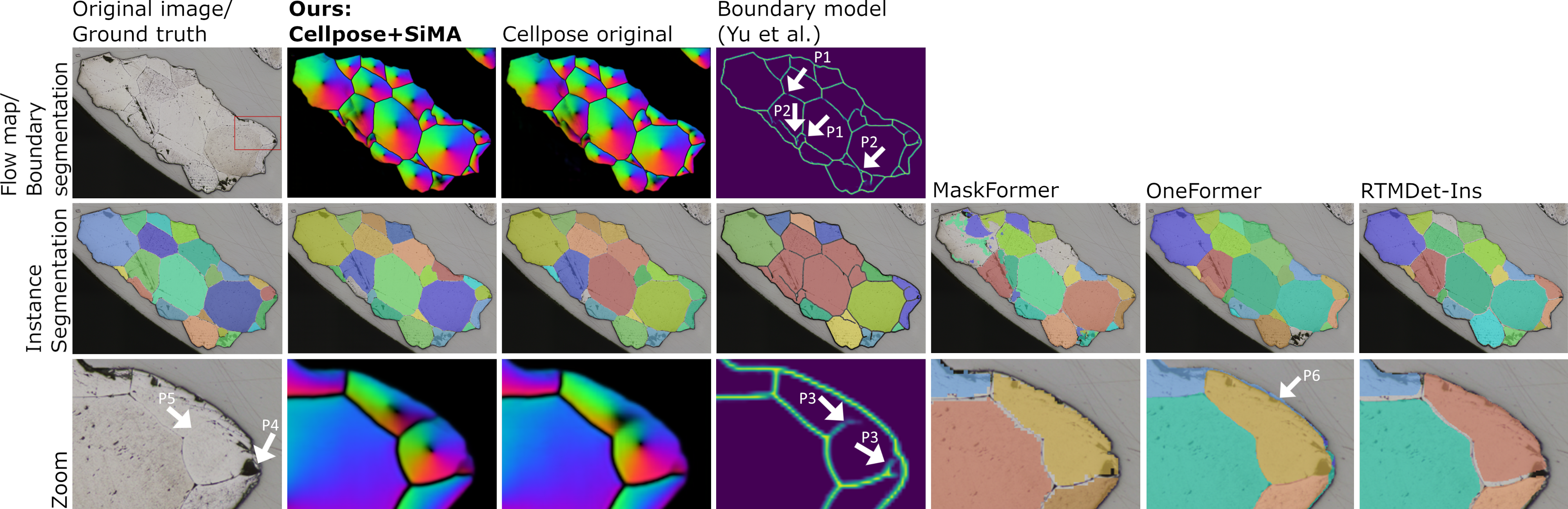}
	\caption{Visual comparison of different model results. The grain consists of several crystals with similar colors, separated in some cases by thin and barely visible boundaries. Thick black scratches additionally complicate the segmentation task. 
	Yu et al.'s boundary model has difficulty in segmenting the boundary, resulting in merged crystals (P1, P3), e.g., the red and blue instance results. It also mistakenly produces two tiny crystals (P2) at the boundary junctions. 
	In the zoomed region, the boundary model is confused by the black spot (P4) located directly at the boundary. The instance segmentation methods perform better in this region, but only our Cellpose+SiMA approach can correctly detect the thin boundary (P5) between the two tiny crystals.  A drawback of MaskFormer, OneFormer, and RTMDet-Ins is that they omit areas where the model is uncertain. OneFormer also has the issue of adding thin segments at the borders of other instances (P6).
 }
	\label{fig5}
\end{figure*}

\subsection{Discussion}
\noindent We showed that each of our multi-scale attention models from Table \ref{tab:result_merging} outperforms the Cellpose original approach. The best PQ and measurement scores are achieved with 4 different scales, referred to as Cellpose+SiMA in the result section.
We also demonstrated that the fusion strategy of multi-scale predictions is crucial, as simple average fusion is consistently outperformed by our attention fusion method and Cellpose original. 
For images with small crystals, Cellpose original can handle them sufficiently, through only a single resizing step. 
However, when the crystals are larger than the input size of the model or vary greatly in size, our multi-scale attention approach outperforms the single resizing step of Cellpose original.

We use the PQ score to measure the predicted segmentation quality, which is related to the accuracy in crystal size measurement (cf. Figure \ref{fig_plot}).
Unlike the PQ score, which takes into account the accuracy of the instance region, the clDice score only measures the accuracy of the boundary, which can be misleading if the region accuracy is low. Therefore, we recommend using the PQ score as a metric for instance segmentation, especially when the goal is to measure the sizes of the segmented objects.

In visually assessing Yu et al.'s boundary model, we observe that the model has difficulties segmenting the boundary in image regions where the crystal boundary is not clearly visible, such as vanishing boundaries, boundaries distorted by scratches, or other occlusions that obscure the boundary. 
This can either lead to gaps in the boundary predictions, resulting in merged instance segments. This can also lead to multiple (incomplete) boundary solutions being predicted, resulting in additional small instance segments.

Instead of using boundary segmentations, our method segments objects by focusing on crystal regions. This allows us to handle the uncertainties at the boundaries better and avoid incorrect segments that are split or merged due to poor boundary predictions. This way, we can handle the uncertainties at the boundaries better and mostly classify them as background pixels, as shown in Figure \ref{fig5}.
Therefore, our instance-based segmentation method is more robust to small segmentation errors and results in fewer incorrectly merged or split segments than boundary-based methods.

Since the boundary model is applied to the image at the original resolution, the downscaling problem mentioned above does not occur here. On the downside, however, this may cause the boundary model to lack global context for large crystals. Their approach does not include image resizing or multi-scale fusion. We assume that our average multi-scale fusion approach is inappropriate for a boundary segmentation model, because boundaries with different widths from different scales could either miss small crystals or mistakenly add small crystals, if the boundaries do not overlap perfectly.

We showed that our multi-scale fusion approach works on our refractory raw material dataset, but we believe that attention fusion of multi-scale predictions is also beneficial for other datasets with high-resolution images that require multiple image resolutions to detect object instances with large differences in size.

\textbf{Limitations:}
From the experiments, we learned that Cellpose has a major limitation; it can only detect the center of the object if the object is fully visible in the input patch. Otherwise, resulting flow maps will be distorted when patches are stitched together. Therefore, appropriate image resizing is key for good segmentation results.
However, this resizing can also cause problems if it is too aggressive and loses important details such as thin boundaries.
For images with thin boundaries and extremely large crystals, Cellpose is not capable of producing correct flow maps. 
This type of data is also challenging for our size-aware attention model SiMA. A possible solution would be to increase the input size of both models, Cellpose and SiMA, to reduce the amount of resizing, but this comes with the price of larger models resulting in higher resource requirements and computational costs.  
This implies, that the image size has an upper limit in our use case - it is a trade-off between how much downsampling is needed and how much downsampling can be done. The downsampling capability is affected by the crystal size, crystal and grain appearance, boundary thickness, or important small details that have to be preserved.

\section{Conclusion}
\noindent
We presented Cellpose+SiMA, an instance segmentation model for high-resolution images with instance objects of large size variation. 
We evaluated our method on a challenging dataset of high-resolution micrographs of refractory  raw material crystals, which have a polycrystalline structure and large variations in crystal size.
We showed that image resizing and multi-scale fusion are crucial for instance segmentation, when objects with different size levels are densely packed together. 
We demonstrated superior performance of our novel size-aware attention model over SOTA boundary and instance segmentation methods.
Our method can effectively fuse instance segmentations from multiple image resolutions resulting in more accurate crystal size measurement.

\bibliographystyle{IEEEtran}
\bibliography{mylibrary}

\end{document}